\documentclass[sigconf]{acmart}

\AtBeginDocument{%
  }

\copyrightyear{2022}
\acmYear{2022}
\setcopyright{rightsretained}

\acmConference[CIKM '22] {Proceedings of the 31st ACM International Conference on Information and Knowledge Management}{October 17--21, 2022}{Atlanta, GA, USA.}
\acmBooktitle{Proceedings of the 31st ACM International Conference on Information and Knowledge Management (CIKM '22), October 17--21, 2022, Atlanta, GA, USA}
\acmISBN{978-1-4503-9236-5/22/10}
\acmDOI{10.1145/3511808.3557599}

\usepackage{subcaption}
\usepackage{tabularx}
\usepackage{multirow}
\usepackage{arydshln}
\usepackage[whole]{bxcjkjatype}

\begin{document}

\title{Expressions Causing Differences in Emotion Recognition \\ in Social Networking Service Documents}


\author{Tsubasa Nakagawa}
\affiliation{%
  \institution{Hosei University}
  \city{Tokyo}
  \country{Japan}
}
\email{tsubasa.nakagawa.5p@stu.hosei.ac.jp}

\author{Shunsuke Kitada}
\affiliation{%
  \institution{Hosei University}
  \city{Tokyo}
  \country{Japan}
}
\email{shunsuke.kitada.8y@stu.hosei.ac.jp}

\author{Hitoshi Iyatomi}
\affiliation{%
  \institution{Hosei University}
  \city{Tokyo}
  \country{Japan}
}
\email{iyatomi@hosei.ac.jp}

\renewcommand{\shortauthors}{Tsubasa Nakagawa, Shunsuke Kitada, \& Hitoshi Iyatomi}

\begin{abstract}
  It is often difficult to correctly infer a writer's emotion from text exchanged online, and differences in recognition between writers and readers can be problematic.
In this paper, we propose a new framework for detecting sentences that create differences in emotion recognition between the writer and the reader and for detecting the kinds of expressions that cause such differences.
The proposed framework consists of a bidirectional encoder representations from transformers (BERT)-based detector that detects sentences causing differences in emotion recognition and an analysis that acquires expressions that characteristically appear in such sentences.
The detector, based on a Japanese SNS-document dataset with emotion labels annotated by both the writer and three readers of the social networking service (SNS) documents, detected ``hidden-anger sentences'' with AUC = 0.772; these sentences gave rise to differences in the recognition of anger.
Because SNS documents contain many sentences whose meaning is extremely difficult to interpret, by analyzing the sentences detected by this detector, we obtained several expressions that appear characteristically in hidden-anger sentences.
The detected sentences and expressions do not convey anger explicitly, and it is difficult to infer the writer's anger, but if the implicit anger is pointed out, it becomes possible to guess why the writer is angry.
Put into practical use, this framework would likely have the ability to mitigate problems based on misunderstandings.

\end{abstract}

\ccsdesc[500]{Information systems~Sentiment analysis}

\keywords{natural language processing, sentiment analysis, text mining}

\maketitle

\section{Introduction}
With the increase in online communication, especially through social networking service (SNS), discrepancies between writers' and readers' perceptions of texts have become an issue~\cite{Yang2009WriterMR}.
Sentiment analysis~\cite{Liu2012SentimentAA} is one practical solution to this problem.
In recent years, researchers have proposed several methods based on machine learning (ML) for classifying the polarity of sentiment (i.e., positive or negative)~\cite{Socher2013RecursiveDM} or for estimating sentiment in terms of type and intensity~\cite{Bostan2018AnAO}.
As types of emotion, Ekman's six emotions (joy, sadness, surprise, anger, fear, and disgust)~\cite{Ekman1992AnAF} and Plutchik's eight emotions (joy, sadness, anticipation, surprise, anger, fear, disgust, and trust)~\cite{Plutchik1980AGP} have been widely used in research.
Emotion estimation from sentences has been implemented in various applications, including an emotion-aware dialogue system~\cite{Firdaus2020EmoSenGS} and  opinion extraction~\cite{Fang2015SentimentAU}.
A support system based on such emotion estimation techniques that presents possible differences in perception between writers and readers is expected to mitigate problems in online communication.

For document data, including publicly available datasets, most of the gold-standard labels used in conventional studies have been assigned by third-party readers~\cite{Aman2007IdentifyingEO,Strapparava2007SemEval2007T1,Mohammad2017EmotionII,Mohammad2018UnderstandingEA,Bostan2020GoodNewsEveryoneAC}.
Therefore, in sentiment analysis, differences in sentiment perceptions between writers and readers cannot be ignored, and the validity of labels is questionable.
Recently, studies have reported quantitative clues about the emotional differences between writers and readers of documents~\cite{Buechel2017ReadersVW}.

\citet{Kajiwara2021WRIMEAN} published WRIME, a Japanese SNS-document dataset that includes eight emotional label information by both the writer and three readers in each data.
They quantitatively compared the emotional intensity of writers and readers and concluded the readers tended to underestimate the writers' emotions, especially anger and trust.
However, they discussed only directly observable differences and did not offer an analysis of more implicit phenomena, such as the causes of underestimation.
We maintain that to estimate the emotions in question more accurately, it is important to identify expressions that cause differences in emotion recognition between writers and readers.

In this study, we propose a new framework using bidirectional encoder representations from transformers (BERT)~\cite{Devlin2019BERTPO} to predict sentences that will cause substantial differences in emotion recognition between writers and readers, and we identified the expressions that may be responsible for these differences.
Our preliminary analysis of this dataset showed that of the eight emotions, anger was the most strongly differentiated in emotion recognition; this paper therefore focuses on this emotion.
To the best of the authors' knowledge, this study is the first attempt to quantitatively analyze clues about differences in emotion recognition between writers and readers of texts used in network communication.
Our model is able to detect sentences that obscured the writer's anger, that is, sentences from which the reader did not infer the writer's anger.
The expressions this study identifies as hidden-anger expressions can be difficult to notice as such, because they lack an explicitly angry meaning.
However, many of them make sense when the anger is pointed out.
Although an angry writer may use such expressions unconsciously, it is important to examine them.

\section{Related Work}
In recent years, pretrained language models such as BERT~\cite{Devlin2019BERTPO} have achieved high performance in sentiment analysis by constructing contextualized representation.
As the performance of emotion estimation has improved, attempts have been made to extract word-level or clause-level causes of emotions in text.
Lee et al. and Xia et al. presented the task of emotion cause extraction~\cite{Chen2010EmotionCD} and emotion-cause pair extraction~\cite{Xia2019EmotionCausePE}, respectively, both using corpus annotated emotions and corresponding causes.

These studies have improved the performance of emotion recognition methods.
However, as Wilson et al. and Parrott pointed out, a single text may contain multiple opinions and emotions~\cite{Wilson2004JustHM}, and human emotions are complex~\cite{Parrott2001EmotionsIS}, so the writer and reader of a text can exhibit differences in emotion recognition.
\citet{Tang2011EmotionMF} attempted to model these differences using a classifier, but the dataset they used lacks labels applied by the writers themselves.
In this paper, we attempt to detect expressions that may cause these differences from sentences annotated by both writers and readers.

\section{Dataset}
This study analyzes the differences in emotion recognition between writers and readers using the WRIME dataset~\cite{Kajiwara2021WRIMEAN}.\footnote{\url{https://github.com/ids-cv/wrime}}
This is a Japanese sentiment analysis dataset.
A total of 43,200 sentences posted on SNS were annotated using four levels of intensity (0-3) of eight different emotions~\cite{Plutchik1980AGP} by the writers (posters) and by three third-party readers.

\begin{figure}[t]
    \begin{minipage}{\linewidth}
        \centering
        \includegraphics[width=0.9\linewidth]{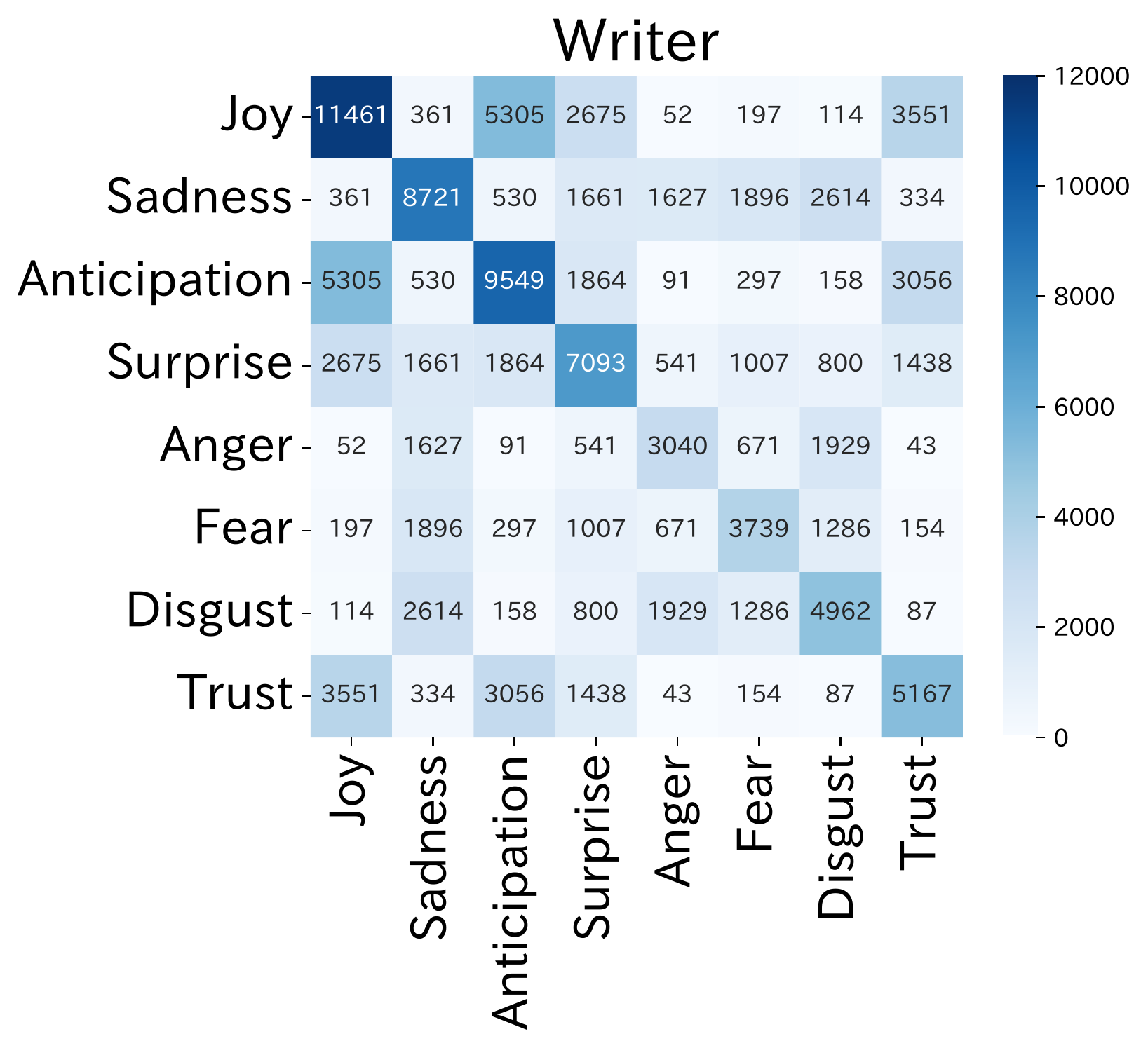}
        \subcaption{Co-occurrence matrix of emotional labels applied by writers.}
        \label{fig:co-occurrence_matrix_writer}
    \end{minipage}
    \begin{minipage}{\linewidth}
        \centering
        \includegraphics[width=0.9\linewidth]{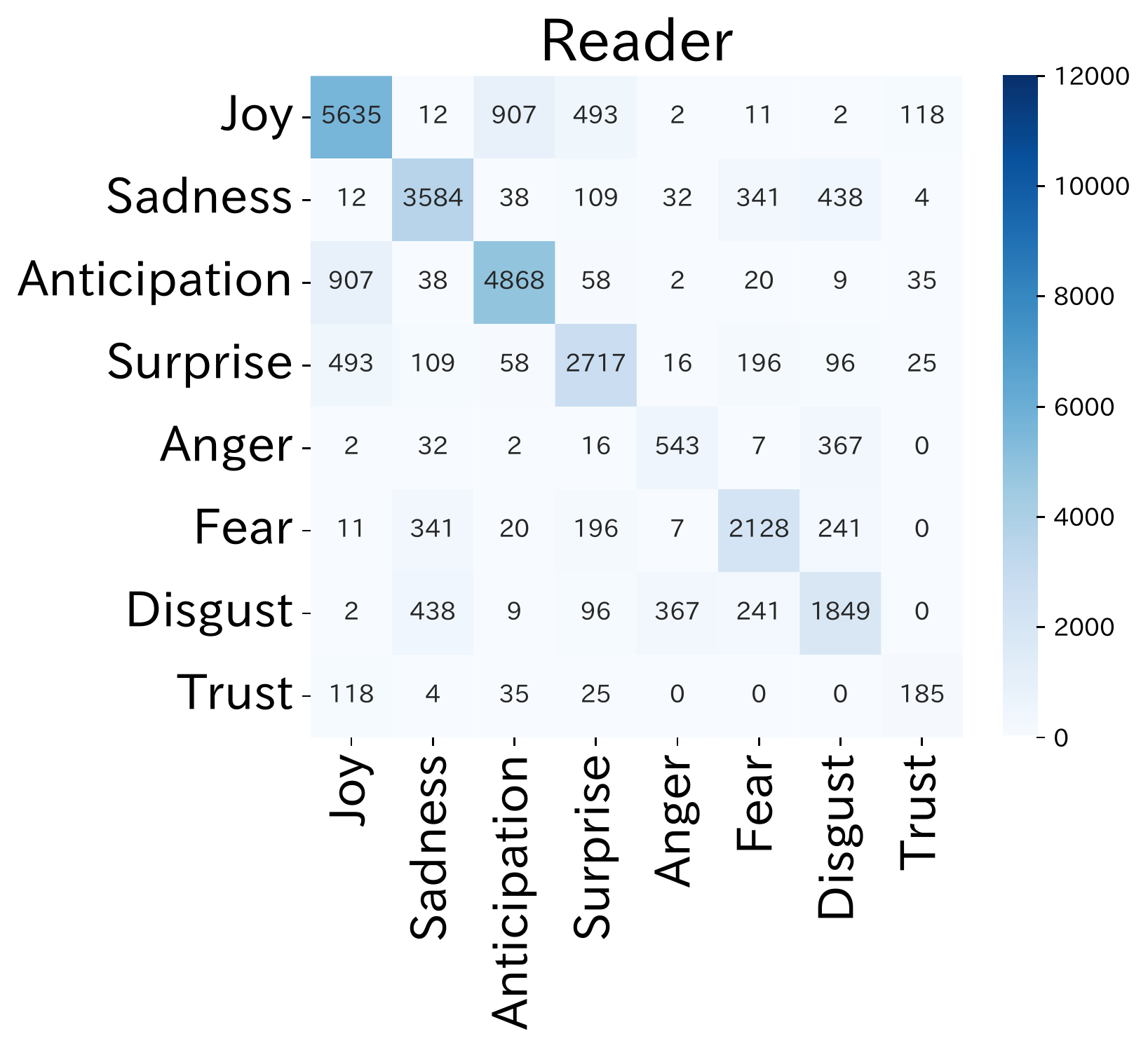}
        \subcaption{Co-occurrence matrix of emotional labels applied by readers.}
        \label{fig:co-occurrence_matrix_reader}
    \end{minipage}
    \caption{Co-occurrence matrix of emotional labels applied by writers and readers.}
    \label{fig:co-occurrence_matrix}
\end{figure}

Figure~\ref{fig:co-occurrence_matrix} shows co-occurrence matrices of emotional labels with intensity of two or more in four levels (0-3) by writers and readers.
The reader's emotional intensity is the average of the emotional labels applied by the three readers.
The matrices show that although writers tend to have multiple emotions, readers tend to perceive only a single emotion.
The matrices also indicate that readers tend to underestimate writers' emotions.
This is especially true for the emotions of anger and trust.
Although the writers applied 3,040 and 5,167 strong anger and trust labels, the readers applied only 543 and 185, respectively.

\section{Proposed Framework}
\begin{table*}[t]
\centering
\caption{Examples of sentences for which sentiment analysis is extremely difficult.}
\label{tab:noisy_sentence_examples}
\begin{tabularx}{0.8\linewidth}{@{}cllll@{}}
\toprule
\multirow{2}{*}{Text} & \multicolumn{4}{X}{雨の日1日1回は見るよねこれ} \\
                      & \multicolumn{4}{X}{\textit{I see this at least once a day on rainy days.}} \\ \hdashline
Anger                 & Writer: 3 & Reader1: 0 & Reader2: 0 & Reader3: 1 \\ \midrule
\multirow{2}{*}{Text} & \multicolumn{4}{X}{あーーー} \\
                      & \multicolumn{4}{X}{\textit{Aaaaah!}} \\ \hdashline
Anger                 & Writer: 2 & Reader1: 0 & Reader2: 0 & Reader3: 0 \\ \midrule
\multirow{2}{*}{Text} & \multicolumn{4}{X}{アッヒョヒョ！！！ファ????????！！！！！！！！！！！！！みたいな鳴き声しとる} \\
                      & \multicolumn{4}{X}{\textit{It's making a chirping sound like "AHHHHHHHHH!!!! Fa????????!!!!!!!!!!!!".}} \\ \hdashline
Anger                 & Writer: 3 & Reader1: 0 & Reader2: 0 & Reader3: 0 \\ \bottomrule
\end{tabularx}
\end{table*}

Based on our analysis of the dataset, we propose a new framework that predicts sentences with substantial differences in emotion recognition and identifies the kinds  of expressions that trigger such differences.
This paper focuses on the emotion of anger for the reason outlined above.
We define sentences in which the writer's anger intensity exceeded the reader's perceived intensity by two or more of the four levels (0-3) as hidden-anger sentences, in which it is difficult for third parties to estimate the writer's anger.
It was possible that the reader's intensity may exceed the writer's, but since it was virtually absent from the current dataset, we did not consider it in this paper.

Here, SNS documents contain not a few of the sentences shown in Table 1, in which it is nearly impossible for models or humans to estimate emotions and sometimes even to understand basic meaning.
In the investigation of what expressions of hidden-anger are being brought about, such sentences are noise and inhibit accurate analysis.
Therefore, our proposed framework consists of two stages; (i) prediction of hidden-anger sentences and, (ii) detection of hidden-anger expressions.
In the first stage, a BERT-based detector is constructed to detect hidden-anger sentences; in the subsequent stage, hidden-anger expressions are detected by analyzing only those sentences the detector has correctly predicted.

\subsection{Prediction of hidden-anger sentences}
In this stage, in order to eliminate noisy sentences and isolate only those sentences that make some sense, we fine-tuned a pre-trained BERT on Japanese Wikipedia\footnote{\url{https://huggingface.co/cl-tohoku/bert-base-japanese-whole-word-masking}} to predict hidden-anger sentences.
Only sentences detected in this stage were used to extract expressions that provide the basis for hidden-anger in the later stage.
We randomly split the dataset into 4:1 for training and evaluation and performed binary classification to detect hidden-anger sentences.
We used cross-entropy as a loss function, a batch size of 32, a dropout rate of 0.1, a learning rate of 2e-5, and Adam~\cite{Kingma2015AdamAM} for optimization.
We stopped the training after three epochs.

\subsection{Detection of hidden-anger expressions}
In this stage, hidden-anger expressions were detected by analyzing the sentences obtained by the BERT-based detector.
We extracted the top 10 words with the largest difference in frequency of occurrence between hidden-anger sentences and other sentences.

\section{Results and Discussion}
\subsection{Prediction of hidden-anger expressions}

\begin{figure}[t]
\centering
\includegraphics[width=0.6\linewidth]{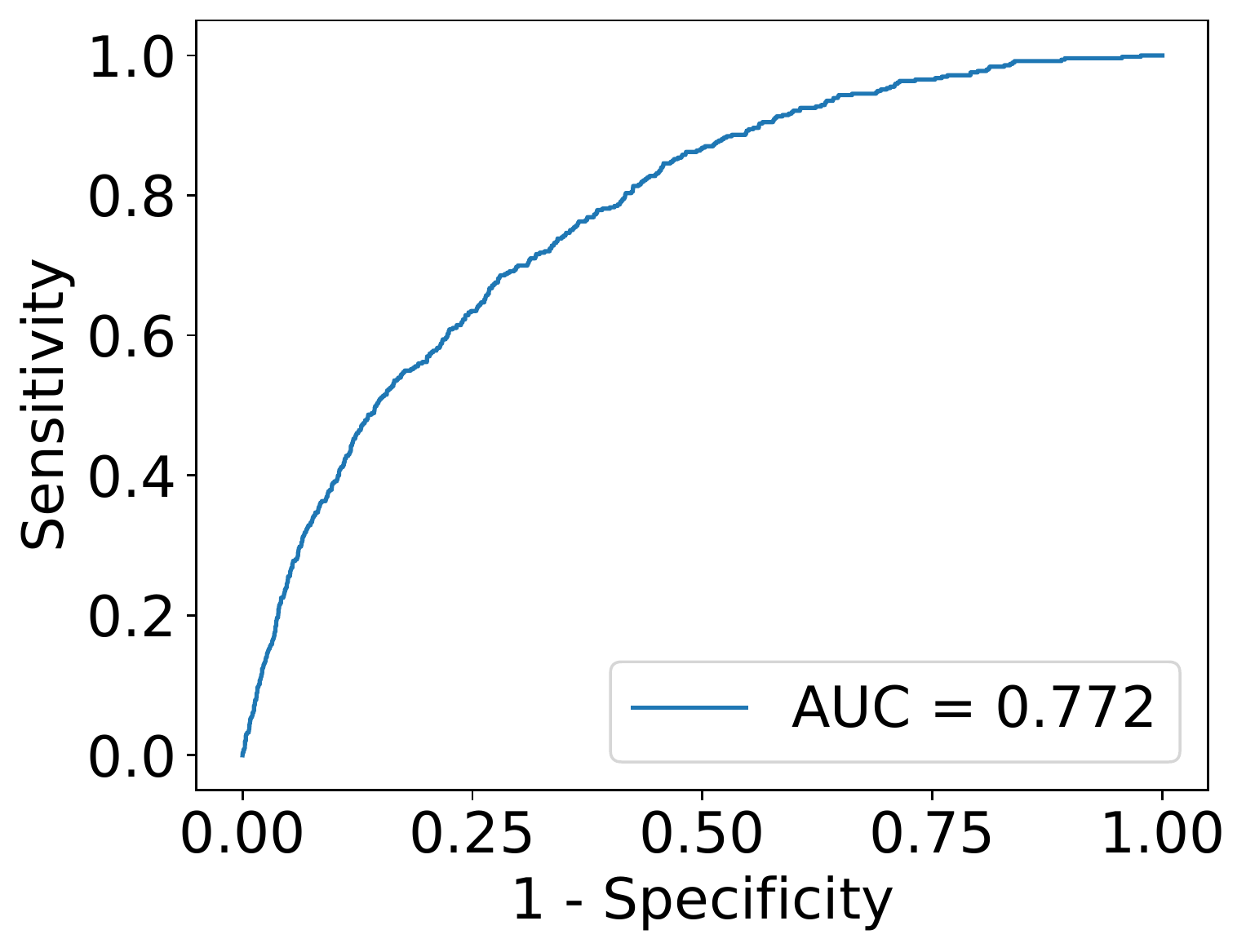}
\caption{ROC curve for detecting hidden-anger sentences.}
\label{fig:roc_curve}
\end{figure}

Figure~\ref{fig:roc_curve} shows the receiver operating characteristic (ROC) curve of the binary classification for detecting hidden-anger sentences.
Our model exhibits clear effectiveness as a detector of hidden-anger sentences.
Of the 493 hidden-anger sentences in the test data, 314 were correctly predicted, and many of these could be read as potentially angry by the writer if read carefully.

\subsection{Detection of hidden-anger expressions}

\begin{figure}[t]
    \begin{minipage}{\linewidth}
        \centering
        \includegraphics[width=0.8\linewidth]{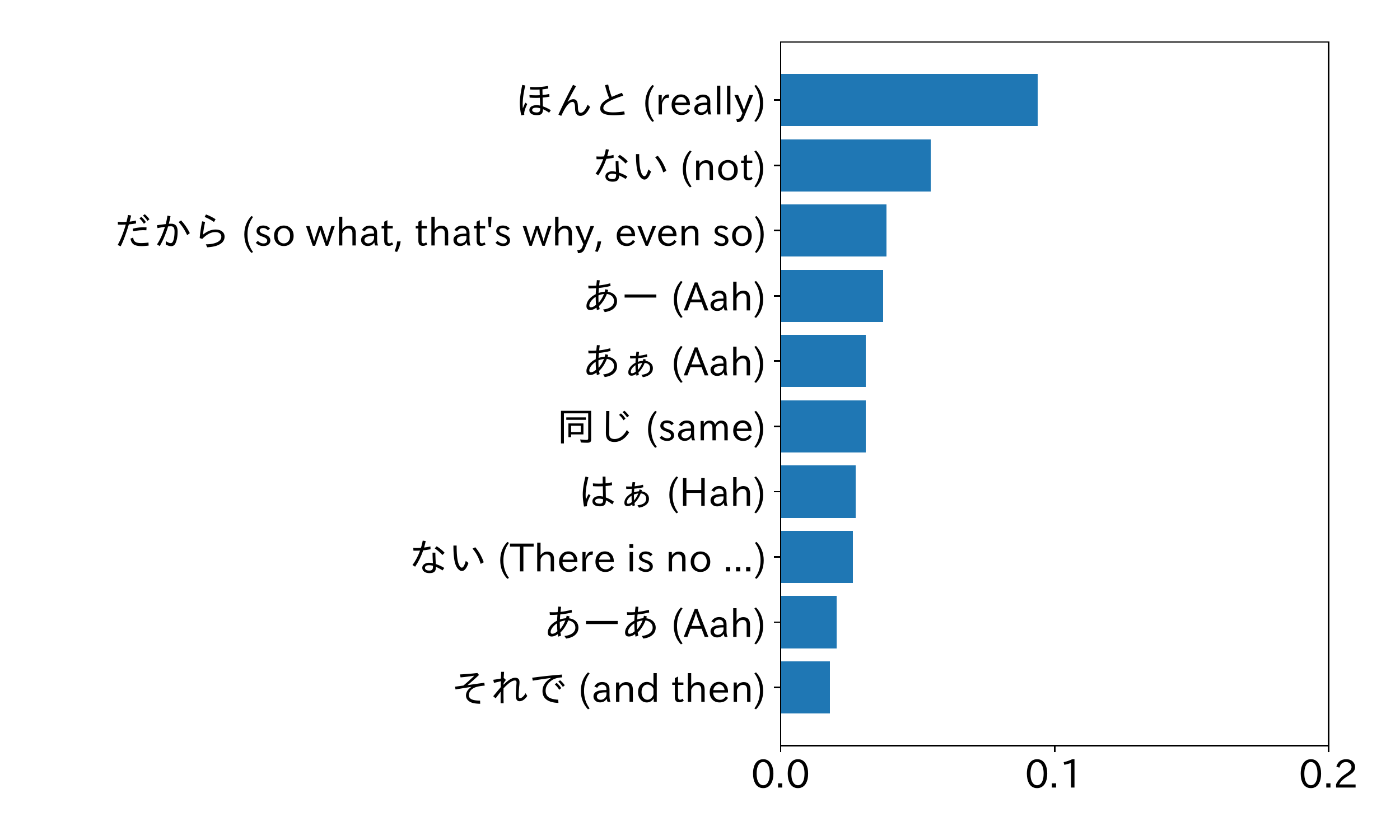}
        \subcaption{Detection using only the labels in the dataset.}
        \label{fig:detected_words_without_bert}
    \end{minipage}
    \begin{minipage}{\linewidth}
        \centering
        \includegraphics[width=0.8\linewidth]{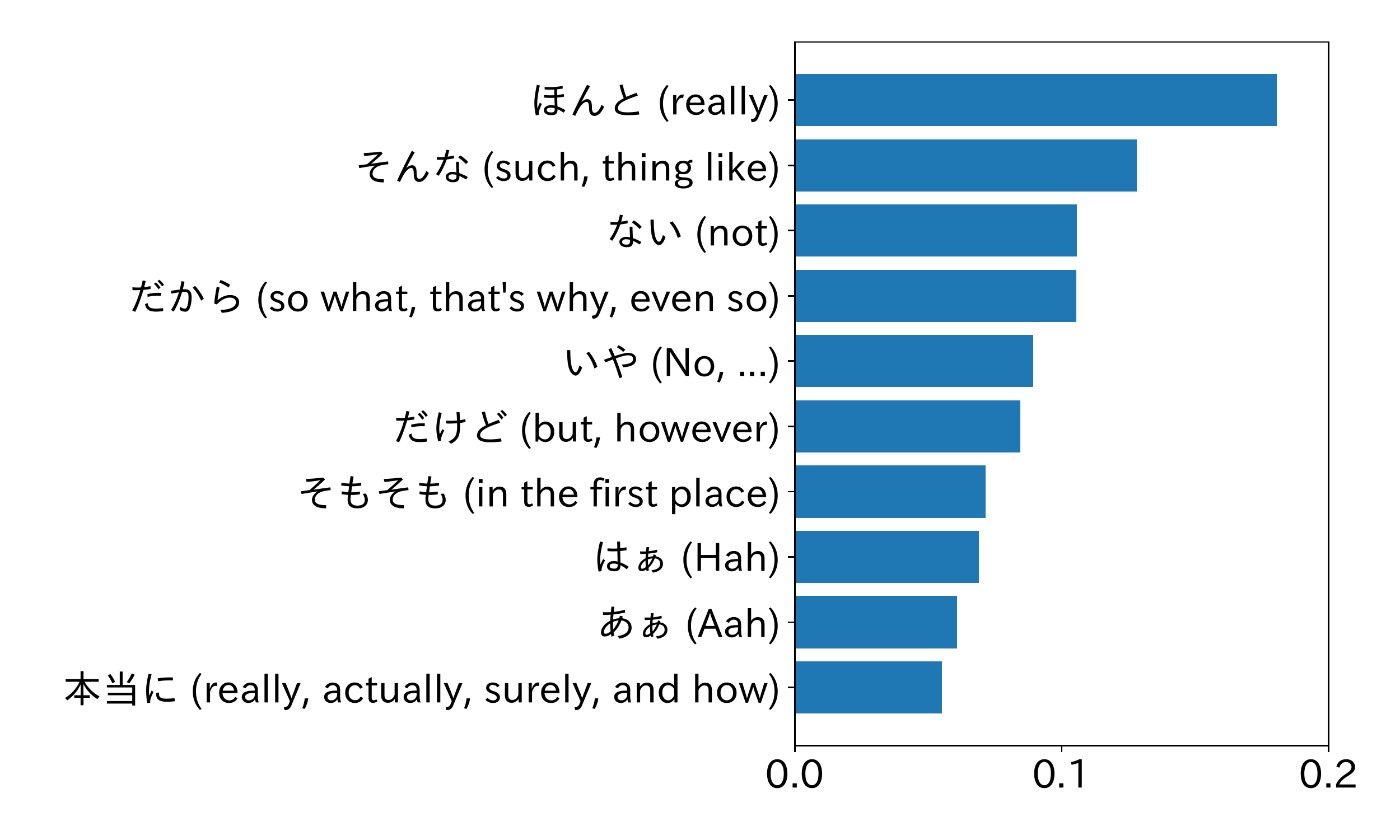}
        \subcaption{Detection using the proposed framework.}
        \label{fig:detected_words_with_bert}
    \end{minipage}
    \caption{Detected hidden-anger expressions and the differences in their frequency between hidden-anger sentences and other sentences.}
    \label{fig:detected_words}
\end{figure}
\begin{table*}[t]
\centering
\caption{Anger-intensity averages for sentences containing the detected words.}
\label{tab:average_anger}
\begin{tabular}{@{}lrrrrr@{}}
\toprule
\multicolumn{1}{c}{Word} & \multicolumn{1}{c}{Writer} & \multicolumn{1}{c}{Reader1} & \multicolumn{1}{c}{Reader2} & \multicolumn{1}{c}{Reader3} & \multicolumn{1}{c}{Writer - Avg. Reader} \\ \midrule
ほんと (really)                       & 0.731 & 0.161 & 0.218 & 0.207 & 0.535 \\
そんな (such, thing like)             & 0.415 & 0.071 & 0.162 & 0.099 & 0.304 \\
ない(not)                             & 0.351 & 0.083 & 0.092 & 0.087 & 0.264 \\
だから (so what, that's why, even so) & 0.587 & 0.198 & 0.215 & 0.157 & 0.397 \\
いや (No, ...)                        & 0.365 & 0.066 & 0.131 & 0.109 & 0.263 \\
だけど (but, however)                 & 0.789 & 0.211 & 0.105 & 0.000 & 0.684 \\
そもそも (in the first place)         & 0.476 & 0.202 & 0.202 & 0.833 & 0.313 \\ \midrule
All sentences                         & 0.234 & 0.047 & 0.057 & 0.051 & 0.182 \\ \bottomrule
\end{tabular}
\end{table*}
\begin{table*}[t]
\centering
\caption{Examples of hidden-anger sentences containing the detected words.}
\label{tab:sentence_examples}
\begin{tabularx}{\linewidth}{@{}cllll@{}}
\toprule
\multirow{2}{*}{Text} & \multicolumn{4}{X}{草取りと朝マラソンと持久走大会だけは\textcolor{red}{ほんと}解せなかった。} \\
                      & \multicolumn{4}{X}{\textit{I \textcolor{red}{really} didn't want to do the weeding, the morning marathon, and the endurance running competition.}} \\ \hdashline
Anger                 & Writer: 3 & Reader1: 0 & Reader2: 0 & Reader3: 0 \\ \midrule
\multirow{2}{*}{Text} & \multicolumn{4}{X}{マックでハンバーガーを注文したら店員に「ハンバーガーは無いです」と言われたので、\textcolor{red}{そんな}わけ無いでしょと思いながらもチーズバーガーを注文した} \\
                      & \multicolumn{4}{X}{\textit{When I ordered a hamburger at McDonald's, the staff told me that they didn't have hamburgers, so I ordered a cheeseburger instead, even though I thought there was \textcolor{red}{no way} that was possible.}} \\ \hdashline
Anger                 & Writer: 2 & Reader1: 0 & Reader2: 0 & Reader3: 0 \\ \midrule
\multirow{2}{*}{Text} & \multicolumn{4}{X}{私のバッシュが売り切れになったんだってさ…。\textcolor{red}{だから}、勝手に私の分がキャンセルになりやがった} \\
                      & \multicolumn{4}{X}{\textit{Apparently, the basketball shoes I ordered were out of stock... And \textcolor{red}{so} my order was canceled without my notice.}} \\ \hdashline
Anger                 & Writer: 3 & Reader1: 1 & Reader2: 2 & Reader3: 0 \\ \midrule
\multirow{2}{*}{Text} & \multicolumn{4}{X}{\textcolor{red}{いや}、もう狂ってる。} \\
                      & \multicolumn{4}{X}{\textit{\textcolor{red}{No}, that's insane.}} \\ \hdashline
Anger                 & Writer: 2 & Reader1: 0 & Reader2: 0 & Reader3: 0 \\ \midrule
\multirow{2}{*}{Text} & \multicolumn{4}{X}{明日出社\textcolor{red}{だけど}こんな時間まであそんじゃた} \\
                      & \multicolumn{4}{X}{\textit{I have to go to work tomorrow, \textcolor{red}{but} I've been playing until this late.}} \\ \hdashline
Anger                 & Writer: 2 & Reader1: 0 & Reader2: 0 & Reader3: 0 \\ \bottomrule
\end{tabularx}
\end{table*}

Figure~\ref{fig:detected_words} shows the top 10 words that can be considered as evidence of hidden-anger sentences, identified using (\subref{fig:detected_words_without_bert}) only the label information in the dataset (i.e., without filtering) and (\subref{fig:detected_words_with_bert}) our proposed framework, that is with filtering using the BERT-based detector. The horizontal axis is the difference in frequency of occurrence between hidden-anger sentences and other sentences, indicating their level as hidden-anger expressions. With the proposed framework, the differences in their frequency are more clearly shown.

Table~\ref{tab:average_anger} presents the anger-intensity averages for sentences containing the words shown in Figure~\ref{fig:detected_words_with_bert}, that is, the hidden-anger expressions detected by our proposed framework.
The rightmost column shows the difference between the writer and the average of the three readers in the anger-intensity labels.
A larger this value indicates a greater difference between the writer's and readers' perceptions of anger.
The values for sentences containing each detected word are greater than 0.182, the value for all sentences in the dataset.
Although this result alone does not determine that the selected words are responsible for hidden-anger expressions, it does confirm that the detected words have an impact on the perception of anger.

Table~\ref{tab:sentence_examples} presents examples of sentences containing the detected words and the anger-intensities annotated in each sentence.
Because these sentences do not involve explicit expressions of anger, it is difficult for readers to estimate the writer's anger, as the labels applied by the three readers make evident.
However, a careful reading of each sentence, with an awareness of the possibility of hidden-anger due to the proposed method, reveals why the writer may have been angry.
In the sentences shown in Table~\ref{tab:noisy_sentence_examples}, that is extremely difficult.
By removing such sentences using the BERT-based detector and then analyzing them, we can detect hidden-anger expressions that humans can interpret accurately through context-aware analysis.

Although it is difficult to rigorously evaluate the validity of the detected hidden-anger expressions, the quantitative evaluation performed in this experiment suggests that the set of words detected by the proposed framework is valid as expressions that trigger differences in the perception of anger in SNS documents.

\section{Conclusion}
In this study, we proposed a framework for detecting expressions that may cause differences in emotion recognition between writers and readers and attempted to detect hidden-anger expressions.
Preremoval of inappropriate sentences using a BERT-based detector in consideration of context enabled us to detect expressions that could not be detected using only label information.
Although the detected expressions did not convey anger explicitly, the writers may have used them unconsciously.
Sharing such findings may ultimately reduce the frequency of mutual misunderstandings.

\begin{acks}
This work was partially supported by JSPS KAKENHI under Grant 21J14143.
\end{acks}

\bibliographystyle{ACM-Reference-Format}
\bibliography{references}

\end{document}